\pdfoutput=1

\documentclass[11pt]{article}

\usepackage[preprint]{coling}

\usepackage{times}
\usepackage{latexsym}

\usepackage[T1]{fontenc}

\usepackage[utf8]{inputenc}

\usepackage{microtype}

\usepackage{inconsolata}

\usepackage{graphicx}
\usepackage{amsmath}
\usepackage{amssymb}
\usepackage{booktabs}
\usepackage[T1]{fontenc}

\usepackage{array}
\newcolumntype{P}[1]{>{\centering\arraybackslash}p{#1}}

\title{Guiding Vision-Language Model Selection for Visual Question-Answering Across Tasks, Domains, and Knowledge Types}

\author{
\textbf{Neelabh Sinha}$^{1}$, 
\textbf{Vinija Jain}$^{2}$\thanks{Work does not relate to position at Meta.}, 
\textbf{Aman Chadha}$^{3}$\thanks{Work does not relate to position at Amazon.} \\
$^1$Georgia Institute of Technology \quad
$^2$Meta AI\quad
$^3$Amazon GenAI \\
{\tt\small nsinha68@gatech.edu, hi@vinija.ai, hi@aman.ai}
}

\begin{document}

\maketitle
\begin{abstract}
Visual Question-Answering (VQA) has become key to user experience, particularly after improved generalization capabilities of Vision-Language Models (VLMs). But evaluating VLMs for an application requirement using a standardized framework in practical settings is still challenging. This paper aims to solve that using an end-to-end framework. We present \texttt{VQA360} -- a novel dataset derived from established VQA benchmarks, annotated with task types, application domains, and knowledge types, for a comprehensive evaluation. We also introduce \texttt{GoEval}, a multimodal evaluation metric developed using GPT-4o, achieving a correlation factor of 56.71\% with human judgments. Our experiments with state-of-the-art VLMs reveal that no single model excels universally, thus, making a right choice a key design decision. Proprietary models such as Gemini-1.5-Pro and GPT-4o-mini generally outperform others, but open-source models like InternVL-2-8B and CogVLM-2-Llama-3-19B also demonstrate competitive strengths, while providing additional advantages. Our framework can also be extended to other tasks\footnote{Code and dataset can be found in the following link: \href{https://github.com/neelabhsinha/vlm-selection-tasks-domains-knowledge-type}{https://github.com/neelabhsinha/vlm-selection-tasks-domains-knowledge-type}.}.
\end{abstract}



\section{Introduction}
\label{sec:introduction}

\begin{figure}[htbp]
    \centering
    \includegraphics[width=0.9\linewidth]{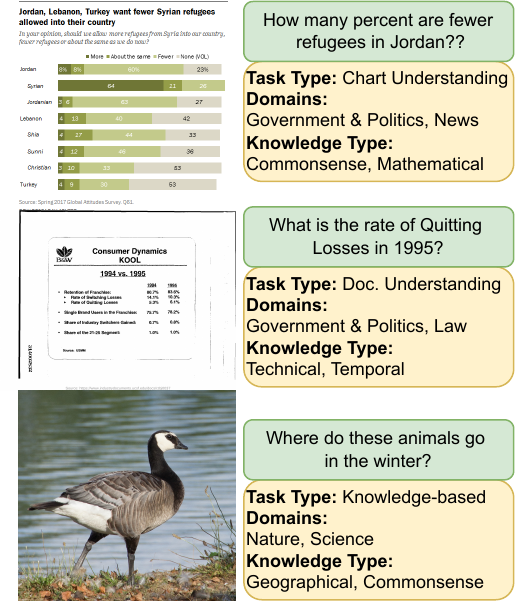}
    \caption{Examples of \texttt{VQA360} tasks and their labels for task types, application domains, and knowledge type in our dataset.}
    \label{fig:data_example}
\end{figure}

Visual Question Answering (VQA)~\cite{vqa_og} is the task of answering a question $q$ about an image $\mathcal{I}$ correctly. This field has been faced with constant challenges in terms of the nature of the problem. For example, the question $q$ can be about the image directly~\cite{vqav2,visual7w,v_in_vqa}, or outside the scope of the image with external knowledge~\cite{okvqa,aokvqa}. The images $\mathcal{I}$ can be a photograph, a mathematical chart~\cite{chartqa,scigraphqa}, a document screenshot~\cite{documentVQA}, or more. 

Dedicated methods~\cite{visual7w,v_in_vqa,analysis_vqa} have long existed to solve different challenges in VQA. But, with the advancement of Vision-Language models (VLMs)~\cite{internvl,Qwen-VL,gemini,gpt4,llava} in multimodal research, several applications have started adapting them due to their versatility. This is because after pre-training~\cite{vila,Wei_2024_CVPR} on vast multimodal datasets~\cite{coco_captions,yfcc100m,conceptual_12m,chartqa,documentVQA,okvqa}, VLMs can effectively generalize across various types of images, and can also incorporate external knowledge beyond the image.

But which VLM to utilize for a given VQA-based requirement? The complexity of this stems from two directions - new VLMs being proposed and the diverse nature of tasks in VQA. New VLMs come up every month now~\cite{llava,llava1_5,Qwen-VL,Qwen2-VL,internvl,far_to_gpt4v,cogvlm,cogvlm2}, and they differ in their architecture, training data, training strategy, size, etc., possessing different capabilites. In addition, users often face technical and business constraints in terms of compute, memory, cost of inference, data, and regulatory risks, which can favor specific VLMs over others. Second, tasks may vary from types such as Chart Question-Answering~\cite{chartqa} or Document Understanding~\cite{documentVQA}, to application domains such as Science, or Sports, and the type of knowledge required, like Geographical Information, Mathematical Reasoning, and beyond. For an application that can fall into one or more such categories, how do you identify the best suited VLM? How to compare them meaningfully? These are gaps in existing literature. Technical reports of VLMs provide benchmarks and comparison, but they are very theoretical and limited.

To bridge the gap in evaluating VLMs on VQA, we propose an end-to-end framework that provides a standardized paradigm for evaluating vision-language models (VLMs) across three key aspects: \textit{task type}, \textit{application domain}, and \textit{knowledge type}. Existing datasets like VQAv2~\cite{vqav2}, OK-VQA~\cite{okvqa}, and ChartQA~\cite{chartqa} offer task instances for training and evaluation but lack labels for other practical aspects. Our framework addresses this by developing and sharing a dataset \texttt{VQA360} derived from the above benchmarks, where tasks are also labeled with their application domains and the knowledge types required for successful completion, as illustrated in Figure~\ref{fig:data_example}, allowing for a evaluation $360^\circ$ in practical settings. Each task can have multiple tags for all these aspects. In addition, traditional NLP evaluation metrics have been shown to be poorly correlated with human judgments for generative models~\cite{evaluating_odqa,g_eval}, an issue that extends to VLM. To address this, we introduce \texttt{GoEval}, a multimodal evaluation metric leveraging GPT-4o~\cite{gpt4}, which demonstrates superior alignment with human judgment compared to existing metrics. Together, they complement each other to provide an end-to-end, completely multimodal evaluation framework. Our framework evaluates 10 variations of 8 VLMs, accommodating diverse requirements such as open-source, resource-efficient, or privacy-compliant models.

In summary, our aim is to address the following \textbf{research questions (RQs)}: \textbf{(1)} How to compare VLMs for different types of VQA tasks in practical settings? \textbf{(2)} How to evaluate those VLM outputs closely with human judgments? \textbf{(3)} As per current SOTA, which VLM is suited for which application, depending on various external constraints?

Our \textbf{key contributions} are as follows:

(i) We release \texttt{VQA360} - a dataset of VQA tasks with three labeled aspects: \textit{task types}, \textit{application domains}, and \textit{knowledge type}, enabling comparison based on different practical requirements.

(ii) We propose \texttt{GoEval}, which is a multimodal evaluation metric based on GPT-4o~\cite{gpt4}, and aligns more closely with human judgments for visual question-answering.

(iii)  We analyze 10 variants of 8 state-of-the-art VLMs of different sizes and families, using our framework to compare their performance.

(iv) Using our analysis, we make recommendations on the best-suited VLMs for a given application requirement under different constraints.
\section{\texttt{VQA360}: A Practical Evaluation Dataset}
\label{sec:exp_dataset}

In this section, we discuss our dataset creation steps we followed in detail, which we propose to utilize for evaluating VLMs for VQA.

\subsection{Source Datasets}
\label{sec:src_datasets}

To be able to evaluate VLMs in a wide variety of QA tasks, \texttt{VQA360} is created from five standard datasets - VQAv2~\cite{vqav2}, OK-VQA~\cite{okvqa}, A-OKVQA~\cite{aokvqa}, ChartQA~\cite{chartqa} and DocumentVQA~\cite{documentVQA}. VQAv2 is an extensive VQA benchmark, while OK-VQA and A-OKVQA are used primarily for knowledge-based VQA, where the answer to the question does not lie within the scope of the image. ChartQA and DocumentVQA are taken to evaluate the performance of VLMs on questions based on mathematical graphs and charts, and documents, respectively. From the test split of each dataset $\mathcal{D}_{test}$, we take $\max(|\mathcal{D}_{test}|, 1145)$ task instances, randomly sampled without replacement. $1145$ was used as it is the minimum size of test set among the five datasets. Thus, our final experimental set contains $5725$ task instances, with equal contributions from each dataset.

\begin{table*}[htbp]
    \centering
    \footnotesize
    \begin{tabular}{l|p{13cm}}
    \toprule
    \textbf{Aspect} & \textbf{Considered Tags} \\
    \midrule
    Application Domains & Anthropology, Formal logic, Economics, History, Law, Government and Politics, Linguistics, Computer Science, Mathematics, Science, Books, Fiction, Movies, News, Reviews, Justice, Professions, Public Places, Knowledge Base, Nature, Nutrition and Food, Social Media, Sports \\
    \hline
    Knowledge Types & Commonsense Knowledge, Visual Knowledge, Cultural Knowledge, Geographical Knowledge, Temporal Knowledge, Social Knowledge, Scientific Knowledge, Technical Knowledge, Mathematical Knowledge, Literary Knowledge, Other \\
    \bottomrule
    \end{tabular}
    \caption{Application domains and knowledge types considered to label all the task instances.}
    \label{tab:valid_aspects}
\end{table*}

\subsection{Instance Tags}

Using the dataset mentioned above only allows us to differentiate based on task types. But there are more ways in which a practical application can be classified. To enable that, we also classify \texttt{VQA360} in two more aspects - application domains, and knowledge types. This is inspired from a recent work~\cite{lm_eval}, but adapted to suit this task. The application domain is the field a task belongs to, such as history, science, sports, etc., and the knowledge type is a specific type of knowledge required, such as geographic, common sense, etc.

\begin{figure*}[htbp]
    \centering
    \includegraphics[width=\linewidth]{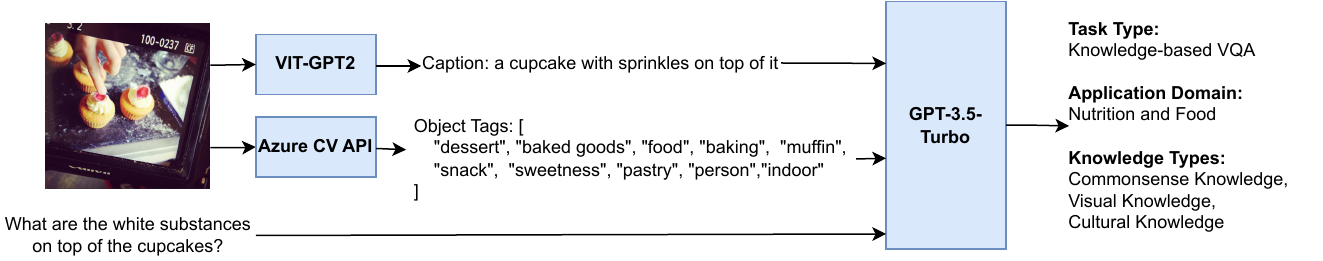}
    \caption{An example with steps taken to generate the instance tags for application domains and knowledge type (task type is mapped directly from the dataset the image is taken).}
    \label{fig:tag_generation_example}
\end{figure*}

Our initial set of tags is crafted manually, with an aim to achieve a broad spectrum of application domains and reasoning types. They are specified in Table~\ref{tab:valid_aspects}, and cover a wide range of domains and knowledge types. We tag each of our task instances in the dataset with one or more application domains and knowledge types using the method described in Figure~\ref{fig:tag_generation_example} and discussed in the following.

\subsection{Generating Instance Tags}

For creating the tags for task types, we map ChartQA to `Chart Understanding', DocumentVQA to `Document Understanding', A-OKVQA and OKVQA to `Knowledge-based Visual Question-Answering' and VQAv2 to `Visual Question-Answering'.

For application domains and knowledge types, we use \texttt{ gpt-3.5-turbo}~\cite{gpt3,gpt35turbo}. To correctly generate instance tags, we require key features of the image and question, and also need to eliminate less useful information from the image to avoid confusions. To achieve this, following a recent work~\cite{generate_then_select}, we generate captions of images from VIT-GPT2\footnote{https://huggingface.co/nlpconnect/vit-gpt2-image-captioning}, and object tags from the Azure Computer Vision API. Using these two as the descriptors of the image and the question, we prompt \texttt{gpt-3.5-turbo} to get the application domains and knowledge type. The prompt used for this task is given in Table~\ref{tab:tags_generation_prompt} in Appendix~\ref{app:appendix}.

After this, we post-process the tags by removing entries that do not belong to any of the entries in Table~\ref{tab:valid_aspects}. If all tags are removed for a task instance, we add "Other" by default. Finally, we manually go through each of the labels and ensure that they are correctly tagged and replace any erroneous tag. The final set contains questions, images, candidate answers, task type, application domains, and knowledge type of all 5725 candidates.

\begin{figure*}[htbp]
    \centering
    \includegraphics[width=0.95\linewidth]{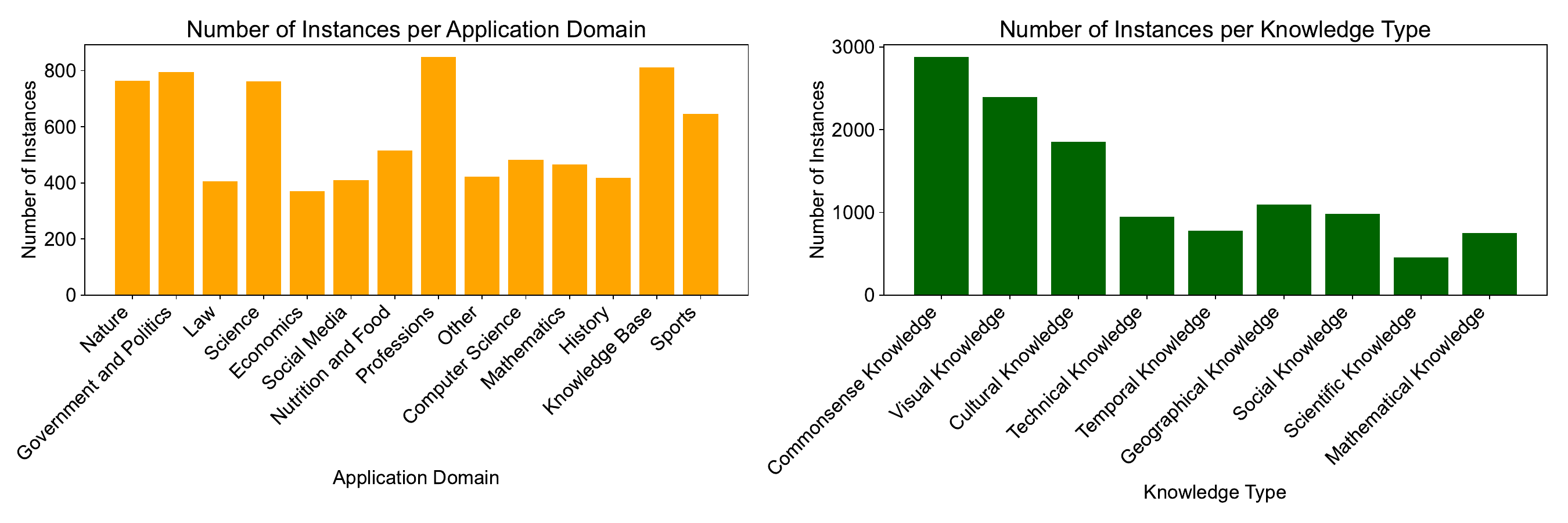}
    \caption{Number of task instances per application domain (left) and knowledge type (right) after generating the instance tags using GPT-3.5-Turbo. All categories are represented by approx 400 instances, which is sufficient for a representative analysis. Categories with $<300$ instances are filtered out, and a task instance can be tagged to multiple categories of a single aspect.}
    \label{fig:tag_statistics}
\end{figure*}

From the instance tags, we remove the tags for which number of instances are less than 300 from our analyses. Please note, we do not remove the task instances entirely, as they may contain other tags included in the study, but just not consider those tags in reporting our results in Section~\ref{sec:vlm_eval} due to less number of instances. This gives a final set of 5 task types, 14 application domains, and 9 knowledge types, which are shown in Figure~\ref{fig:tag_statistics}. We also report statistis of \texttt{VQA360} in Table~\ref{tab:tag_and_descriptor_statistics}.

\begin{table}[htbp]
    \centering
    \footnotesize    
    \begin{tabular}{p{0.4\linewidth}|P{0.12\linewidth}P{0.12\linewidth}P{0.12\linewidth}}
    \toprule
     \textbf{Statistic (per instance)} & \textbf{Mean}  & \textbf{Std.} & \textbf{Max}  \\
     \midrule
      Caption Length   & 46.56 & 8.84 & 96 \\
      Object Tags & 13.01 & 7.83 & 59 \\
      Application Domains & 1.7 & 0.68 & 7 \\
      Knowledge Type & 2.19 & 0.83 & 9 \\
      \bottomrule
    \end{tabular}
    \caption{Average, std. and maximum of length of caption and count of object tags in generated image descriptors, and number of application domains and knowledge types per instance. This clarifies that significant number of task instances have multiple application domains and knowledge type tags.}
    \label{tab:tag_and_descriptor_statistics}
\end{table}

\texttt{VQA360} allows extended analysis of diverse VQA tasks and allows looking into performance of VLMs from entirely new perspectives. Further, the creation methodology can also be extended for enriching other datasets and creating benchmarks for evaluating VLMs in different settings.

We use \texttt{VQA360} for rest of the analysis, and also release it publicly (linked in footnote of Page 1), for the research community to utilize in future research. Although it contains instances of existing benchmarks, it is a one-stop benchmark for an extended evaluation with labels for task types, application domains, and knowledge types. We also provide object tags and captions.

\section{\texttt{GoEval}: A VQA Evaluation Metric}
\label{sec:goeval}

Evaluating QA using lexical matching has significant limitations, particularly when correct answers don't match with set of gold answers~\cite{evaluating_odqa}. A recent work~\cite{evaluating_odqa} evaluated traditional VQA metrics against GPT-based evaluation for open-domain QA, and found it to be more aligned with human judgments. 

Another alternative is to evaluate using NLG metrics such as ROUGE~\cite{rouge}, METEOR~\cite{meteor}, and then use a threshold to determine correct/wrong. However, they also suffer from similar limitations, as evaluated from previous work~\cite{g_eval}. A probable solution is to use BERTScore~\cite{bertscore}, which compares texts in embedding space, focusing more on semantic similarity.

These metrics may be promising, but they are not equipped to handle multimodal settings. To fill this gap, we propose \texttt{GoEval} - a multimodal metric based on GPT-4o, which can be used to evaluate VQA. Similar to existing works~\cite{evaluating_odqa,g_eval}, we create a prompt, and ask GPT-4o if the generated answer is correct. However, to incorporate the vision modality in making judgments, we also use the image. We use zero-shot evaluation with prompting, using both GPT-4o and GPT-4o-mini, to compare and contrast performance v/s cost trade-offs.

Formally, we pick a prompt function $\mathcal{P}$ from the first row of Table~\ref{tab:go_eval_prompts} of the Appendix~\ref{app:appendix}, and generate a prompt $p = \mathcal{P}(q, r, c)$ based on question $q$, the reference answer set $r$ and the candidate response $c$. Using this prompt and the image, we prompt GPT-4o~\cite{gpt4,gpt4o} to ask if it is correct. We also compare our prompt against two other ways -- without using the image, and without reference answers. This is to understand which technique shows highest correlation with human judgments. We don't compare with traditional methods, as it is already done in the above discussed works~\cite{evaluating_odqa,g_eval}, and our method without using the image is similar to what is used in~\cite{evaluating_odqa}.

\subsection{Validating \texttt{GoEval}}

To validate \texttt{GoEval}, we first generate outputs on our experimental dataset using Gemini-1.5-Pro~\cite{gemini}, a state-of-the-art VLM. Then, we manually evaluate all the answers on the validation set, marking $0$ for incorrect answer and $1$ for correct answer. These are the results of human evaluation. 

We compare BERTScore precision, recall, F1-score, and six variants of \texttt{GoEval} using GPT-4o and GPT-4o mini~\cite{gpt4}, with and without reference answers, and with and without using images with human evaluations in terms of accuracy and Kendall Tau. The exact prompts that we use are outlined in Table~\ref{tab:go_eval_prompts} of Appendix~\ref{app:appendix}. When we don't use the image, we also alter the template text by a little to not ask model to refer the image. The results are detailed in Table~\ref{tab:goeval_comparison}.

\begin{table}[htbp]
    \centering
    \footnotesize
    \begin{tabular}{c|cc}
        \toprule
        \textbf{Method} & \textbf{Acc (\%)} & \textbf{$\tau$} \\
        \midrule
        BERTScore-p & 1.30 & 36.21 \\
        BERTScore-r & 1.30 & 12.68 \\
        BERTScore-f1 & 1.30 & 28.50\\
        \midrule
        \texttt{GoEval} (-R, -I) & 49.91 & 16.17 \\
        \texttt{GoEval}-mini (-R, -I) & 52.70 & 10.84 \\
        \midrule
        \texttt{GoEval} (-R) & 71.43 & 35.94 \\
        \texttt{GoEval}-mini (-R) & 64.01 & 24.37 \\
        \midrule
        \texttt{GoEval} & 78.48 & 52.43 \\
        \texttt{GoEval}-mini & \textbf{80.33} & \textbf{56.71} \\
        \bottomrule
    \end{tabular}
    \caption{Comparison of different evaluation methods: Accuracy (Acc.) and correlation ($\tau$ for Kendall's Tau) of evaluation metrics with human judgement. -R indicates absence of reference answers, -I indicates absence of image from the request. \texttt{GoEval} with all components (reference answers, image) performs the best.}
    \label{tab:goeval_comparison}
\end{table}

From the results, we can see that \texttt{GoEval} with reference answer and image provides the best alignment with human judgment. Without using the image (-R, -I), the performance proves to be very weak, depicting that existing metrics that do not utilize the image will perform poorly on VQA. We believe this is because image context is crucial for VQA, and in the absence of it, the model isn't able to reason well whether the provided answer is correct or not. Moreover, the differences between models are largely influenced by the amount of information. For example, performance of \texttt{GoEval} and \texttt{GoEval}-mini with image but without reference only has a difference of $\Delta\tau=11$, but as soon as the reference answers are added, $\tau$ increases by 20 units. This is because the reference answers serve as an extra guidance in addition of the image to determine the correctness of the candidate answer, making the model perform better. 

In summary, \texttt{GoEval} shows high accuracy and Kendall's Tau correlation with human evaluation, and \texttt{GoEval}-mini marginally outperforms \texttt{GoEval}. This complements \texttt{VQA360} to provide an end-to-end cohesive framework for VLM evaluation in entirely multimodal settings. 
\section{Comparative Evaluation of VLMs}
\label{sec:vlm_eval}

We evaluate state-of-the-art (SOTA) VLMs in practical settings using our framework using \texttt{GoEval}-mini, since it demonstrated maximum correlation with human judgments, and is more cost-efficient.

For VLMs, we use InternVL-2 1B and 8B~\cite{internvl,far_to_gpt4v}, PaliGemma-3B~\cite{paligemma}, Qwen-2-VL 2B and 7B~\cite{Qwen-VL,Qwen2-VL}, LlaVa-1.6-Mistral-7B~\cite{llava1_5}, CogVLM2-Llama-3-19B~\cite{cogvlm,cogvlm2}, Gemini-1.5 Flash and Pro~\cite{gemini}, and GPT-4o-Mini~\cite{gpt4}. The rationale behind choosing these models is to have sufficient diversity to allow users to choose the appropriate VLM based on other constraints. 

All models except Gemini-1.5 and GPT-4o are open sourced, which gives freedom to customize the models as desired, and host it in-house. It takes away the privacy and regulatory risk of sending data to a third-party, and reduces operational and opportunity cost factor, as these APIs are costly with rate limits. In-house hosting allows a relatively fixed cost. We have taken smaller models in the 1B-3B range which can be used in resource-constrained environments and on-device AI.

For all open-source models, we use the HuggingFace implementation with the image and question in a prompt recommended by the model card, since we want to evaluate all scenarios uniformly. We use Gemini APIs\footnote{https://ai.google.dev/gemini-api/docs} OpenAI API\footnote{https://platform.openai.com/docs/overview} for Gemini and GPT-4o-mini. The results are discussed in the following subsections. More details of the artifacts used are given in Table~\ref{tab:artifacts} of the Appendix~\ref{app:implementation_details}.

\subsection{Correlation Between VLM Outputs}
\label{sec:correlation_vlm}

\begin{figure}[htbp]
    \centering
    \includegraphics[width=\linewidth]{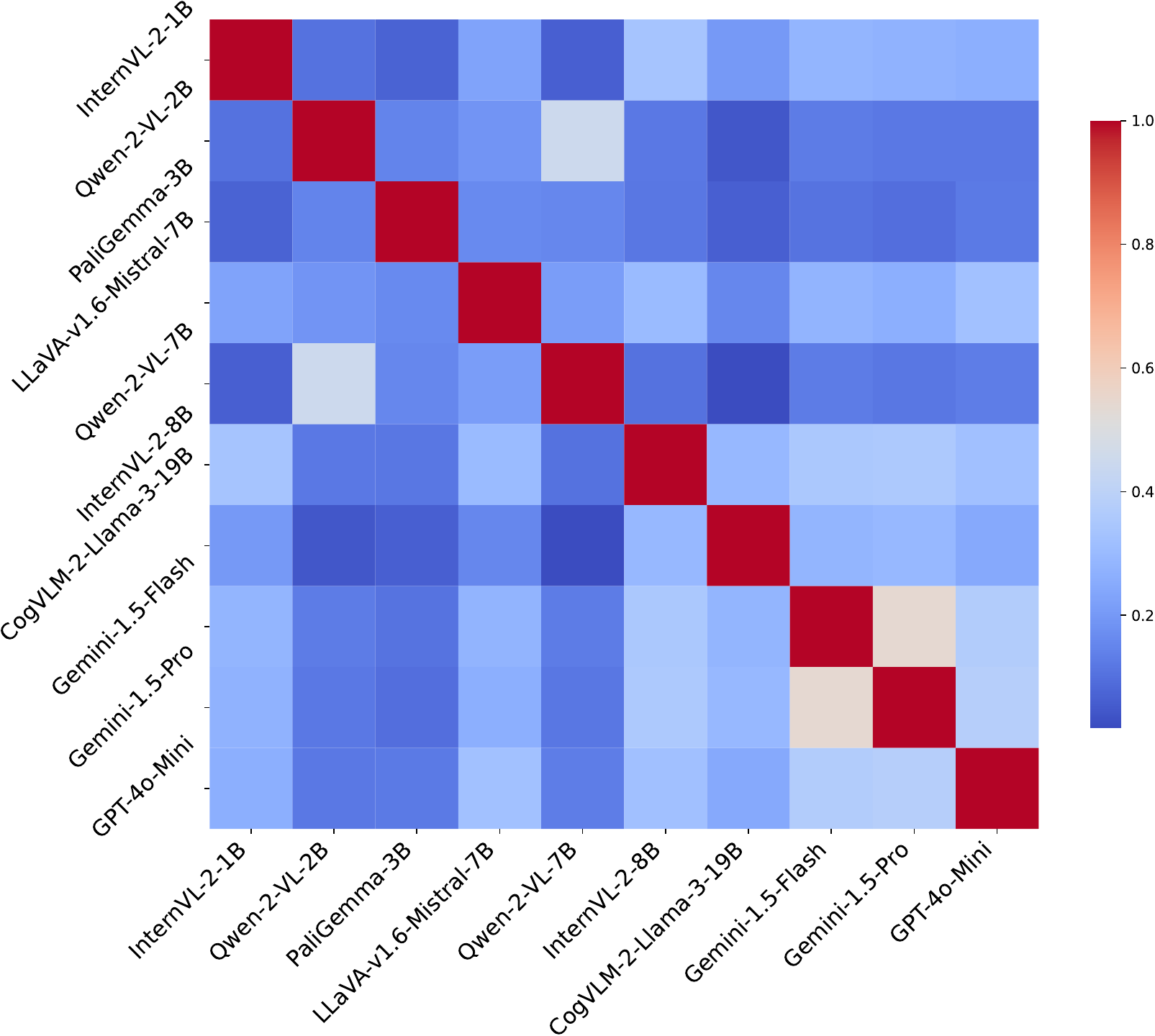}
    \caption{Correlation of \texttt{GoEval}-mini values between performance of different VLMs for all task instances. The low correlation values for outputs between all models indicate different VLMs perform differently with task instances.}
    \label{fig:vlm_performance_correlation}
\end{figure}

Our first hypothesis was that all VLMs do not perform similarly with all task instances. They have their own strengths and weaknesses. To establish that, we evaluated the correlation between the outputs of different models for all tasks and document the results in Figure~\ref{fig:vlm_performance_correlation}.

From the figure, most of the correlations are low. This shows that the performance difference between VLMs is not just in terms of a global statistic, but also differs for individual task instances. Thus, some VLMs that might be great in a subset of tasks may be poor in another subset. 

The highest correlation is observed between Gemini-1.5 Flash and Pro, followed by Qwen-2-VL 2B and 7B, and InternVL-2-1B and 8B. Since they are from the same family, they might have similar training data, training strategy, and architecture. Some open-source VLMs also exhibit higher correlations (light blue) with Gemini and GPT, though these correlations remain relatively low.

Since task types, application domains, and knowledge types are key factors in which tasks can be differentiated in practical settings, we move to analyzing the performance on those factors.

\subsection{Evaluation on Task Types}
\label{sec:comparison_task_types}

We evaluated VLMs on four different task types: chart understanding using ChartQA dataset~\cite{chartqa}, Document Understanding using DocVQA dataset~\cite{documentVQA}, knowledge-based VQA using A-OKVQA~\cite{aokvqa} and OK-VQA~\cite{okvqa}, and general VQA using VQAv2~\cite{vqav2}. The performance of VLMs across these is summarized in Table~\ref{tab:comparison_task_types}.

\begin{table*}[htbp]
\centering
\footnotesize
\begin{tabular}{p{2.2cm}|p{0.9cm}p{0.9cm}p{0.9cm}p{0.9cm}p{0.9cm}p{0.9cm}p{0.9cm}p{0.9cm}p{0.9cm}p{0.9cm}}
\toprule
\textbf{Task Type} & \textbf{Intern-VL-2-1B} & \textbf{Qwen-2-VL-2B} & \textbf{Pali-Gemma-3B} & \textbf{LLaVA-1.6-Mistral-7B} & \textbf{Qwen-2-VL-7B} & \textbf{Intern-VL-2-8B} & \textbf{Cog-VLM-2-Llama-3-19B} & \textbf{Gemini-1.5-Flash} & \textbf{Gemini-1.5-Pro} & \textbf{GPT-4o-Mini} \\
\midrule
Chart U. & 46.81 & 12.31 & 12.84 & 22.88 & 14.06 & 65.68 & 67.34 & 62.17 & \textbf{67.66} & 60.36 \\
Document U. & 45.68 & 15.81 & 12.49 & 30.07 & 20.00 & 65.76 & 68.17 & 64.36 & \textbf{70.53} & 54.68 \\
KBQA & 45.41 & 19.83 & 19.22 & 58.65 & 26.94 & 71.43 & 62.77 & 70.66 & 77.31 & \textbf{84.04} \\
VQA & 55.81 & 26.29 & 23.58 & 57.64 & 29.52 & 68.65 & 57.18 & 69.11 & 73.25 & \textbf{77.74} \\
\bottomrule
\end{tabular}
\caption{Mean \texttt{GoEval}-mini scores for different task types for all VLMs. \textbf{Bold} numbers indicate best results. Gemini-1.5-Pro performs better in chart and document understanding, while GPT-4o-mini performs better in the other two (U. = Understanding, KBQA = Knowledge-based VQA).}
\label{tab:comparison_task_types}
\end{table*}

From the table, the closed models, which are believed to be SOTA, outperform the smaller, open-sourced models. This is expected considering that those models are larger. Within them, we identify that Gemini-1.5-Pro performs significantly better than GPT-4o-mini when extracting information from an image is key, like interpreting documents and charts. On the other hand, where knowledge and comprehension are critical, GPT-4o-mini outperforms Gemini-1.5-Pro. If processing cost is a factor, Gemini-1.5-Flash can be chosen with approximately 7-10\% performance loss. 

Among open-source models, InternVL-2 shows promising results with both 1B and 8B models given their size, and can be chosen if open-source models are needed. CogVLM-2-Llama-3-19B also competes closely with Gemini-1.5-Flash in Chart and Document understanding tasks. Llava-1.6-Mistral-7B performs acceptable in knowledge-based VQA and VQA, but the performance degrades drastically in the other two categories, where visual comprehension is critical, exposing its limitations in that area. Qwen-2-VL variants and PaliGemma-3B surprisingly prove to be weak in all tasks. 

\subsection{Evaluation on Application Domains}
\label{sec:comparison_app_domains}

\begin{figure}[htbp]
    \centering
    \includegraphics[width=\linewidth]{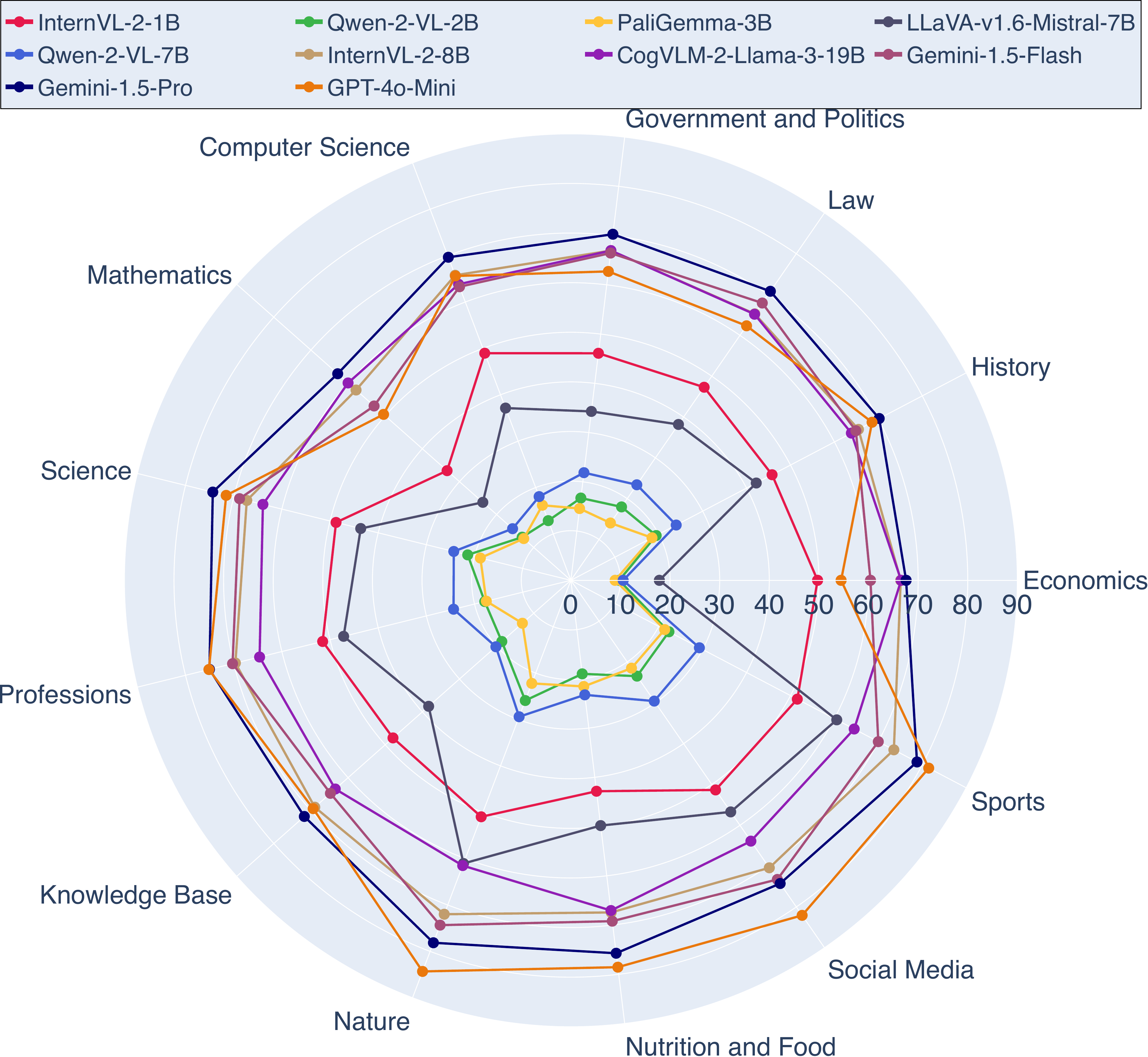}
    \caption{Mean \texttt{GoEval}-mini scores for different application domains for all VLMs. Gemini-1.5-Pro and GPT-4o-mini are the best performing closed models, with CogVLM-2-LlaMa-3-19B and InternVL-2-8B performing the best amongst open models.}
    \label{fig:comparison_domains}
\end{figure}

We evaluated the VLMs in all application domains that had more than 300 task instances, and the results are shown in Figure~\ref{fig:comparison_domains}.

If the VLMs were similar in all application domains, their result would be perfectly circular. However, we see that most of the VLM performance graphs have aberrations in multiple categories, highlighting variance in strengths. In most cases, the weakness of one of the few VLMs is compensated for by the strength of others. So, choosing a VLM wisely according to the application need can help mitigate weaknesses.

Among the closed models, GPT-4o-mini proves to be the best in four categories - Nature, Nutrition and Food, Social Media and Sports, and Gemini-1.5-Pro proves to be the best in all other categories. GPT-4o-mini doesn't even remain second best in some categories like Mathematics, Economics, Law, and is outperformed by many open-sourced models here. Gemini-1.5-Flash remains strong with a performance deficit compared to Pro, but in Social Media tasks, it almost matches Pro. Therefore, while the appropriate model should be selected based on requirements, the Gemini-1.5-Pro generally looks to be the best overall choice.

In the open-source model category, InternVL-2-8B and CogVLM-2-Llama-3-19B are the best possible choices. CogVLM-2-Llama-3-19B generally performs well in more academic topics like Mathematics, Computer Science, law, Government and Politics, but suffers a lot of performance degradation in more social topics like Nature, Nutrition and Food, Social Media, Sports. InternVL-2-8B also shows similar traits, but the difference is relatively less. For academic topics, these models even outperform some of the closed models. Llava-1.6-Mistral-7B is one model that shows exactly opposite trait than this, being limited in academic topics as compared to social topics. Qwen-2 variants and PaliGemma show weak results in all domains, like in task types. InternVL-2-1B remains the best choice if a small model is required, with decent results using 1B parameters.

\subsection{Evaluation on Knowledge Types}
\label{sec:comparison_knowledge_types}

\begin{figure}[htbp]
    \centering
    \includegraphics[width=\linewidth]{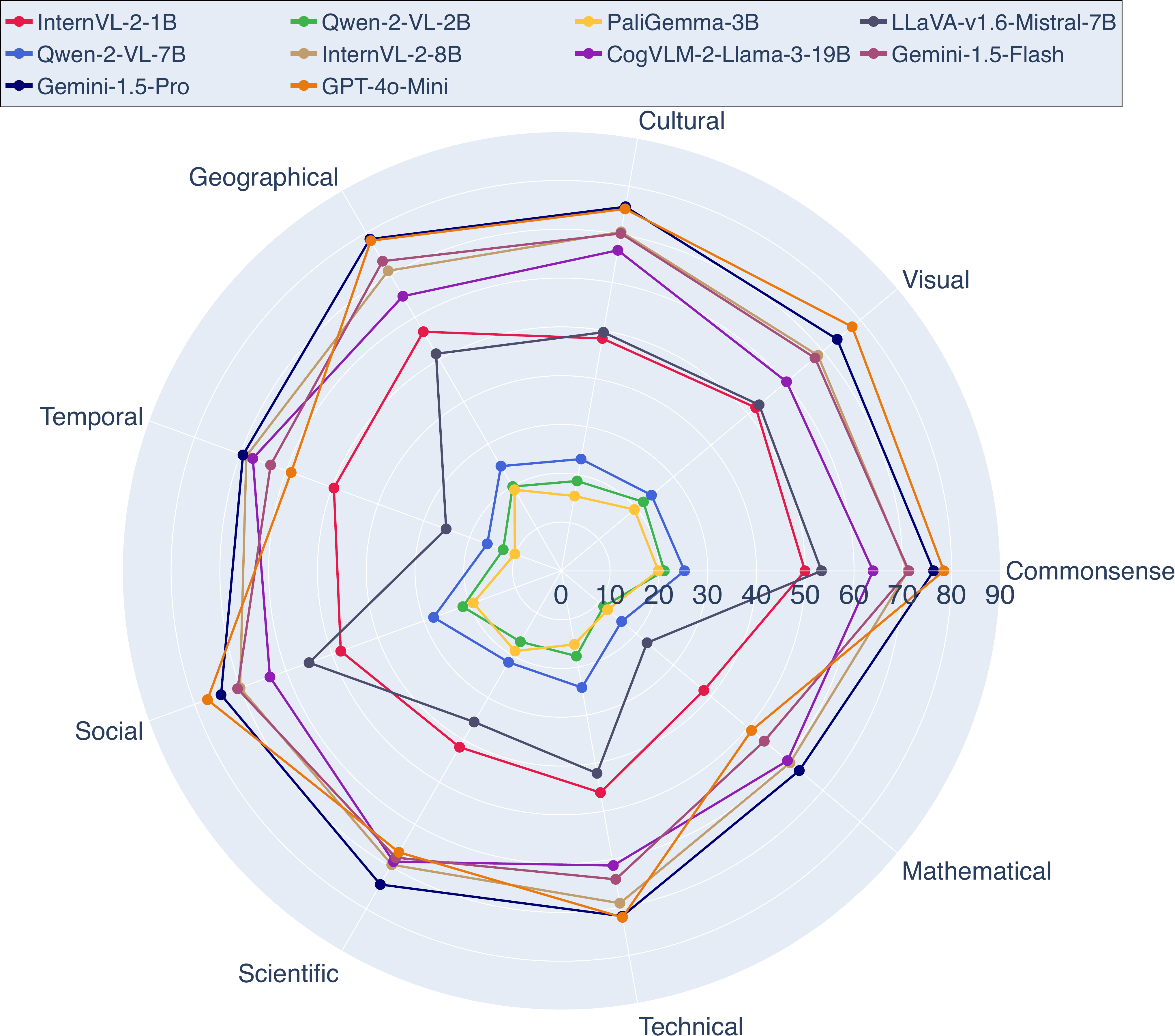}
    \caption{Mean \texttt{GoEval}-mini scores for different knowledge types for all VLMs. Gemini-1.5-Pro and GPT-4o-mini perform the best in most knowledge types, while InternVL-2 models also demonstrate competitive performance based on their size.}
    \label{fig:comparison_knowledge_types}
\end{figure}

Similar to application domains, we evaluate all VLMs on all knowledge types where number of task instances is greater than 300. We demonstrate the results using a similar radar chart in Figure~\ref{fig:comparison_knowledge_types}.

In the closed models category, the SOTA performance is again shared by Gemini-1.5-Pro and GPT-4o-mini. But unlike in application domains, GPT-4o-mini competes more strongly with Gemini-1.5-Pro on several knowledge types. Visual, Social and Commonsense knowledge is where the advantage of GPT-4o-mini over Gemini-1.5-Pro is maximum, again depicting its strength in social knowledge types. It is still considerably weak in temporal, scientific, and mathematical knowledge, falling behind even some of the small open-source models. Gemini-1.5-Pro will still be the best overall choice, but different models can be selected based on the results obtained for specific knowledge types. Gemini-1.5-Flash again follows similar trend, with slight performance difference from Pro.

Comparing open-source models, InternVL-2-8B and CogVLM-2-Llama-3-8B are the best choices here as well, but here we can see that InternVL-2-8B outperforms the latter in all knowledge types. InternVL-2-1B continues to be the best overall choice if small models are required, and the Qwen-2 variants and PaliGemma continue to be the least effective. Therefore, InternVL-2-8B is the best overall choice for open source models.

\subsection{Overall Analysis and Recommendations}

In subsections~\ref{sec:comparison_task_types},~\ref{sec:comparison_app_domains} and~\ref{sec:comparison_knowledge_types}, we evaluated all vision-language models under three different aspects of task types, application domains and knowledge types. We also discussed their strengths and weaknesses in those categories and recommended different VLMs to use under different requirements. In this section, we will take a higher level look at everything together. 

We identified that Gemini-1.5-Pro and GPT-4o-mini are different. In general, GPT-4o-mini is weaker in analytical tasks, like tasks of Mathematics or Economics domain, or scientific or mathematical knowledge types. It not only falls behind Gemini-1.5-Pro in these tasks, but also behind open models like InternVL-2-8B or CogVLM-2-Llama-3-19B. However, it is strong is social and topical tasks like Nature, Sports, Nutrition and Food domains. Gemini-1.5-Pro proves strong more generally, but is expensive. It is either the best, or comes close second or third best. Therefore, if cost is a significant factor, Gemini-1.5-Flash can be considered as a decent alternative at a performance deficit of around 7-10\%.

InternVL-2-8B and CogVLM-2-Llama-3-19B are strong open-source models. Due to size differences, resource availability also contributes to deciding which model to use. CogVLM-2-Llama-3-19B is better at more academic tasks, that belong to domains like History, Law, Computer Science, etc., or knowledge types like Temporal, Scientific or Mathematical Knowledge. InternVL-2-8B is a more general capable model that demonstrates more suitability in broader application requirements. In some cases, it outperforms GPT-4o-mini as well. Possessing superior performance in addition to other advantages of open-sourced model makes it a strong choice. It can also be aligned for downstream tasks to improve performance.

Among small models suited for on-device AI, resource-efficient environments, InternVL-2-1B proves is strongest overall, significantly outperforming models like Qwen2-VL-2B and PaliGemma-3B in all categories. 

The Qwen-2-VL variants and PaliGemma-3B did not prove fit for use in our experimental settings, being very weak on all categories. LlaVa-1.6-Mistral-7B also performs average, similar to InternVL-2-1B, but is weak in all aspects compared to InternVL-2-8B, a similar-sized model. 

Since Gemini-1.5-Pro was the most successful model, we demonstrate some of the qualitative examples using that model in Table~\ref{tab:qualitative_examples} in the Appendix~\ref{app:appendix}. Finally, Tables~\ref{tab:domain_results_all} and~\ref{tab:knowledge_types_results_all} of the Appendix~\ref{app:appendix} contain quantitative results on all application domains and knowledge types, respectively, including the categories that were excluded from the study of the main paper.
\section{Conclusion}
\label{sec:conclusion}

In this paper, we propose a comprehensive framework for evaluating Vision-Language Models (VLMs) across diverse visual question-answering (VQA) tasks, addressing specific application requirements. Our framework introduces a novel evaluation paradigm that classifies VQA tasks along three dimensions: \textit{task types}, \textit{application domains}, and \textit{knowledge types}. To support this, we release \texttt{VQA360}, a dataset annotated across 4 task types, 22 application domains, and 15 knowledge types, derived from established VQA benchmarks. We also present \texttt{GoEval}, a new evaluation metric to complement it, leveraging GPT-4o to integrate visual and textual information, achieving a 56.71\% correlation with human judgments and outperforming traditional metrics.

Through experiments with 10 state-of-the-art VLMs, we observe significant performance variation across categories, with no single model proving universally optimal. Proprietary models like Gemini-1.5-Pro and GPT-4o-mini achieve the highest overall performance, while open-source models such as InternVL-2-8B and CogVLM-2-Llama-3-19B excel in specific scenarios. Our findings provide actionable insights for task-specific VLM selection, and establishes a evaluation framework that can be extended to other vision-language tasks, fostering progress in multimodal research.

\bibliography{egbib}

\appendix
\section{Appendix}
\label{app:appendix}

In this appendix, we provide additional details and results related to our work. The implementation details can be found in Section~\ref{app:implementation_details}. Table~\ref{tab:tags_generation_prompt} provides prompts designed to classify tasks by domain (e.g., "Anthropology," "Computer Science") and by the type of knowledge required (e.g., "Commonsense Knowledge," "Visual Knowledge"). Table~\ref{tab:go_eval_prompts} presents the prompts used in \texttt{GoEval} to verify whether a candidate answer is correct, with or without reference answers, both for visual and text-only evaluation. Additionally, Table~\ref{tab:qualitative_examples} details some qualitative examples using the best overall model that we found - Gemini-1.5-Pro. Finally, Tables~\ref{tab:domain_results_all} and~\ref{tab:knowledge_types_results_all} contain quantitative results on all application domains and knowledge types, respectively, including the categories that were excluded from the study of the main paper.

\subsection{Implementation Details}
\label{app:implementation_details}

In this subsection, we will discuss more details around implementation. Table~\ref{tab:artifacts} contains all the model cards, which contain exact details of how we implemented all the VLMs, and the recommended prompt templates that we used in our evaluation.

\begin{table}[htbp]
    \centering
    \begin{tabular}{p{0.55\linewidth}|p{0.25\linewidth}}
        \toprule
        \textbf{Artifact} & \textbf{Link}  \\
        \midrule
        VQAv2 & \href{https://huggingface.co/datasets/HuggingFaceM4/VQAv2}{Dataset Card}  \\
        OK-VQA & \href{https://huggingface.co/datasets/HuggingFaceM4/OK-VQA}{Dataset Card}  \\
        A-OKVQA & \href{https://huggingface.co/datasets/HuggingFaceM4/A-OKVQA}{Dataset Card}  \\
        ChartQA & \href{https://huggingface.co/datasets/HuggingFaceM4/ChartQA}{Dataset Card}  \\
        DocumentVQA & \href{https://huggingface.co/datasets/HuggingFaceM4/DocumentVQA}{Dataset Card}  \\ \midrule
        InternVL-2-1B & \href{https://huggingface.co/OpenGVLab/InternVL2-1B}{Model Card} \\
        Qwen-2-VL-2B & \href{https://huggingface.co/Qwen/Qwen2-VL-2B-Instruct}{Model Card} \\
        PaliGemma-3B & \href{https://huggingface.co/google/paligemma-3b-pt-224}{Model Card} \\
        Qwen-2-VL-7B & \href{https://huggingface.co/Qwen/Qwen2-VL-7B-Instruct}{Model Card} \\
        LLaVA-v1.6-Mistral-7B & \href{https://huggingface.co/llava-hf/llava-v1.6-mistral-7b-hf}{Model Card} \\
        InternVL-2-8B & \href{https://huggingface.co/OpenGVLab/InternVL2-8B}{Model Card} \\
        CogVLM-2-Llama-3-19B & \href{https://huggingface.co/THUDM/cogvlm2-llama3-chat-19B}{Model Card} \\
        Gemini-1.5-Flash & \href{https://deepmind.google/technologies/gemini/flash/}{Model Card} \\
        Gemini-1.5-Pro & \href{https://deepmind.google/technologies/gemini/pro/}{Model Card} \\
        GPT-4o-Mini & \href{https://openai.com/index/gpt-4o-mini-advancing-cost-efficient-intelligence/}{Model Card} \\
        \midrule
        BERTScore & \href{https://github.com/Tiiiger/bert_score}{Doc (used using Roberta Large)} \\
        \bottomrule
    \end{tabular}
    \caption{Details of artifacts used with artifact links.}
    \label{tab:artifacts}
\end{table}

For all the models, we prepend the question with a static text saying `Only answer the below question. Do not provide any additional information. In addition, we resize all images to $448\times448$ before sending them through the model. This is the input to every model. For decoding the output, we use Greedy sampling, since fluency is not a key factor in VQA as long as the answers are correct. We use $max\_new\_tokens=2048$ for all models. 

The implementation of executing all models and our evaluation metric can be found in the Code provided. We also provide the implementation of captions and object tags that can be used if this framework is being adapted to other tasks. All the configuration parameters and hardware used are detailed in Table~\ref{tab:hardware_config}.

\begin{table}[htbp]
    \centering
    \begin{tabular}{p{0.4\linewidth}|p{0.5\linewidth}}
        \toprule
        \textbf{Configuration Parameter} & \textbf{Specification} \\
        \midrule
        Number of GPUs & 1, 2 for CogVLM-2 \\
        GPU Model & Nvidia A40 \\
        GPU Memory Capacity & 48 GB \\
        Batch Size & 8 \\
        Image Resolution & $448\times448$ (224 for PaliGemma) \\
        Maximum New Tokens & 2048 \\
        \bottomrule
    \end{tabular}
    \caption{Hardware and model configuration details used in the experiments, highlighting specialized settings for certain models.}
    \label{tab:hardware_config}
\end{table}

We also use 4-bit quantization and Flash Attention 2~\cite{flash_attention_2} wherever supported for memory and execution efficiency.

\subsection{Using this Work to Select VLM}

The prerequisite to using this work is to lay down the problem statement and its scope along with other system parameters that should include, but should not be limited to resource availability, data availability, system constraints, resource or data processing budget, acceptable performance bounds, etc. 

Start with finding the task type, application domain and reasoning type closest to your requirement from Table~\ref{tab:comparison_task_types},~\ref{tab:domain_results_all} and~\ref{tab:knowledge_types_results_all}. Next, from your design constraints, identify some sets of VLMs acceptable for your solution. For example, if using on-device AI, you might only be able to use either small VLMs, or closed model accessible by APIs, depending on your acceptable performance bounds, and regulatory aspects of being able to share data across, having the inference budget for using APIs, and so on. Refer to Section~\ref{sec:vlm_eval} for a more detailed discussion around which models will fit which needs. Between those, check the performance of the subset of categories and subset of models, and choose the best model.

These models provide a comparison on a uniform foundation so that a comparative analysis can be done. These models can further be customized as desired for best outputs as per other design parameters and needs.

\begin{table*}
\centering
\begin{tabular}{p{3cm}|p{12cm}}
\toprule
\textbf{Aspect} & \textbf{Prompt to Extract Tags} \\
\midrule
Prompt to Extract Application Domains & 
Following are source application domains: \newline
Anthropology, Books, Computer Science, Economics, Fiction, Formal logic, Government and Politics, History, Justice, Knowledge Base, Law, Linguistics, Movies, Mathematics, Nature, News, Nutrition and Food, Professions, Public Places, Reviews, Science, Social Media, Sports. \newline
There is an image which can be described as: \{caption\}. \newline
The image has the following objects: \{object\_tags\}. \newline
A user is asking the following question on the image: \{question\}, \newline
What type of application domain does this task belong to? Choose one or many alternatives from the above options. \newline
Return output as list of strings as JSON Object. Example: \{\{'application\_domain': ['domain\_a', 'domain\_b']\}\} \\
\midrule
Prompt to Extract Knowledge Type & 
Following are the names and explanation of types of knowledge: \newline
Commonsense Knowledge: Knowledge about the world that humans learn from their everyday experiences (e.g., many donuts being made in a cart implies they are for sale rather than for personal consumption). \newline
Visual Knowledge: Knowledge of concepts represented visually (e.g., muted color palettes are associated with the 1950s). \newline
Cultural Knowledge: Understanding cultural references, norms, and practices (e.g., knowing that a red envelope is associated with good luck in Chinese culture). \newline
Temporal Knowledge: Awareness of historical events, timelines, and changes over time (e.g., recognizing a specific style of clothing as being from the 1980s). \newline
Geographical Knowledge: Information about locations, landmarks, and regional characteristics (e.g., identifying a famous monument like the Eiffel Tower in Paris). \newline
Social Knowledge: Understanding social interactions, relationships, and behaviors (e.g., recognizing that a handshake is a form of greeting). \newline
Scientific Knowledge: Knowledge from various scientific domains like physics, biology, chemistry, astronomy, etc. (e.g., understanding that certain plants are poisonous). \newline
Technical Knowledge: Familiarity with technology, machinery, and tools (e.g., identifying parts of a computer or types of construction equipment). \newline
Mathematical Knowledge: Basic mathematical concepts and their applications (e.g., understanding geometric shapes or calculating areas). \newline
Literary Knowledge: Awareness of literature, authors, and genres (e.g., recognizing characters from classic novels). \newline
There is an image which can be described as: \{caption\}. \newline
The image has the following objects: \{object\_tags\}. \newline
A user is asking the following question on the image: \{question\}. \newline
What type of knowledge is required to answer the question? Choose one or many alternatives from the above options. \newline
Return output as list of strings as JSON Object. Example: \{\{'knowledge\_type': ['knowledge\_a', 'knowledge\_b']\}\} \\
\hline
\end{tabular}
    \caption{Prompts used to generate domain and knowledge type tags using the question, image caption and object tags.}
    \label{tab:tags_generation_prompt}
\end{table*}

\begin{table*}
    \centering
    \begin{tabular}{c|c|c}
        \toprule
        \textbf{Prompt} & \textbf{Reference} & \textbf{Image} \\
        \midrule
        \begin{tabular}{p{0.6\textwidth}}
        Question: \{question\} \\
        Reference Answers: \{reference\} \\
        Candidate Answer: \{candidate\} \\
        \\
        Consider Reference Answers to be multiple answers provided for the given question in context with the above image. If there are multiple answers, they are separated by semi-colon(;). Based on the image, is the candidate answer a correct answer for the given question? Answer only `yes' if the candidate answer is correct or only `no' if it is not.
        \end{tabular} & 
        \checkmark & \checkmark \\
        \midrule
        \begin{tabular}{p{0.6\textwidth}}
        Question: \{question\} \\
        Candidate Answer: \{candidate\} \\
        \\
        Based on the image, is the candidate answer a correct answer for the given question? Answer only `yes' if the candidate answer is correct or only `no' if it is not.
        \end{tabular} & 
        \texttimes & \checkmark \\
        \midrule
        \begin{tabular}{p{0.6\textwidth}}
        Question: \{question\} \\
        Reference Answers: \{reference\} \\
        Candidate Answer: \{candidate\} \\
        \\
        Consider Reference Answers to be multiple answers provided for the given question in context. If there are multiple answers, they are separated by semi-colon(;). Based on the context, is the candidate answer a correct answer for the given question? Answer only `yes' if the candidate answer is correct or only `no' if it is not.
        \end{tabular} & 
        \checkmark & \texttimes \\
        \midrule
        \begin{tabular}{p{0.6\textwidth}}
        Question: \{question\} \\
        Candidate Answer: \{candidate\} \\
        \\
        Based on the context, is the candidate answer a correct answer for the given question? Answer only `yes' if the candidate answer is correct or only `no' if it is not.
        \end{tabular} & 
        \texttimes & \texttimes \\
        \bottomrule
    \end{tabular}
    \caption{Prompts used with GoEval in different settings. (\checkmark)~represents included, (\texttimes)~represents excluded.}
    \label{tab:go_eval_prompts}
\end{table*}

\begin{table*}[htbp]
    \centering
    \begin{tabular}{p{0.2\textwidth}|p{0.15\textwidth}|P{0.15\textwidth}|P{0.1\textwidth}|P{0.1\textwidth}|P{0.1\textwidth}}
    \toprule
       \textbf{Image}  & \textbf{Question} & \textbf{Labels} & \textbf{Response} & \textbf{Human Evaluated Label} & \texttt{GoEval}-mini  \\
    \midrule
        \raisebox{-\totalheight}{\includegraphics[width=0.2\textwidth]{assets/results/qualitative_examples/A-OKVQA_931.png}} & The cat is doing what? & laying down, sleeping, laying down, laying, sleeping, resting, laying, laying, laying down, resting & Lying down & \checkmark & \checkmark \\
    \midrule
        \raisebox{-\totalheight}{\includegraphics[width=0.2\textwidth]{assets/results/qualitative_examples/A-OKVQA_379.png}} & What number is closest to the number of people that are pushing the bus? & eight, eight, eight, ten, twelve, nine, twenty, ten, nine, 15 & 10 & \texttimes & \checkmark \\
    \midrule
        \raisebox{-\totalheight}{\includegraphics[width=0.2\textwidth]{assets/results/qualitative_examples/ChartQA_1084.png}} & Is there any difference between the male and female values? & No & No & \checkmark & \texttimes \\
    \midrule
        \raisebox{-\totalheight}{\includegraphics[width=0.2\textwidth]{assets/results/qualitative_examples/DocumentVQA_421.png}} & What is the name of the company mentioned in logo? & golden tobacco limited, Golden Tobacco Limited & The logo & \texttimes & \texttimes \\
    \midrule
         \raisebox{-\totalheight}{\includegraphics[width=0.2\textwidth]{assets/results/qualitative_examples/VQAv2_144330.png}} & What is the message written on? & sign, sign, metal sign, stop sign, signs, street sign, sign, metal sign, street sign & A sign & \checkmark & \checkmark \\
    \bottomrule
    \end{tabular}
    \caption{Some qualitative examples with Gemini-1.5-Pro. (\checkmark)~represents correct answer, saying that response correctly answers the question, (\texttimes)~represents incorrect answer.}
    \label{tab:qualitative_examples}
\end{table*}

\begin{table*}[htbp]
\centering
\small
\begin{tabular}{p{3.2cm}|p{0.8cm}p{0.8cm}p{0.8cm}p{0.8cm}p{0.8cm}p{0.8cm}p{0.8cm}p{0.8cm}p{0.8cm}p{0.8cm}}
\toprule
\textbf{Domain} & \textbf{Intern-VL-2-1B} & \textbf{Qwen-2-VL-2B} & \textbf{Pali-Gemma-3B} & \textbf{LLaVA-v1.6-Mistral-7B} & \textbf{Qwen-2-VL-7B} & \textbf{Intern-VL-2-8B} & \textbf{Cog-VLM-2-Llama-3-19B} & \textbf{Gemini-1.5-Flash} & \textbf{Gemini-1.5-Pro} & \textbf{GPT-4o-Mini} \\
\midrule
Anthropology & 46.47 & 20.33 & 16.60 & 60.58 & 21.99 & 73.86 & 63.18 & 71.91 & 74.69 & \textbf{82.16} \\
Books & 55.56 & 21.37 & 10.68 & 40.60 & 20.94 & 70.51 & 66.23 & 63.79 & \textbf{72.22} & 66.95 \\
Computer Science & 48.96 & 12.86 & 16.18 & 37.14 & 18.05 & 65.77 & 63.83 & 63.29 & \textbf{69.65} & 65.61 \\
Economics & 49.73 & 9.73 & 8.92 & 17.84 & 10.54 & 66.49 & 66.58 & 60.39 & \textbf{67.57} & 54.47 \\
Fiction & 53.70 & 22.22 & 24.07 & 61.11 & 29.63 & 81.48 & 57.41 & 62.96 & 70.37 & \textbf{83.33} \\
Formal logic & 31.96 & 23.71 & 14.43 & 37.50 & 22.68 & 56.70 & 50.00 & 58.33 & \textbf{63.92} & 58.76 \\
Government and Politics & 46.11 & 16.71 & 14.57 & 34.30 & 21.86 & 66.96 & 66.96 & 66.45 & \textbf{70.28} & 62.72 \\
History & 45.80 & 19.42 & 18.47 & 42.21 & 23.98 & 65.47 & 63.86 & 64.88 & \textbf{70.19} & 68.56 \\
Justice & 41.07 & 23.21 & 10.71 & 46.43 & 32.14 & 66.07 & 62.50 & 64.81 & \textbf{73.21} & 71.43 \\
Knowledge Base & 47.91 & 18.60 & 13.05 & 38.30 & 20.20 & 68.97 & 63.44 & 64.76 & \textbf{71.78} & 69.40 \\
Law & 47.29 & 17.98 & 14.04 & 38.18 & 23.40 & 65.27 & 65.17 & 67.92 & \textbf{70.79} & 62.34 \\
Linguistics & 48.00 & 17.78 & 20.89 & 44.89 & 21.33 & 67.56 & 58.04 & 63.64 & 75.00 & \textbf{79.46} \\
Mathematics & 33.33 & 13.12 & 12.69 & 23.71 & 15.70 & 57.85 & 60.00 & 52.99 & \textbf{62.80} & 50.44 \\
Movies & 53.80 & 22.78 & 20.89 & 51.90 & 32.91 & 69.62 & 63.29 & 66.24 & 76.58 & \textbf{78.06} \\
Nature & 50.98 & 25.92 & 22.22 & 61.05 & 29.41 & 71.99 & 61.50 & 74.34 & 78.14 & \textbf{84.32} \\
News & 54.59 & 17.84 & 11.35 & 31.89 & 17.84 & 68.11 & 68.51 & 70.79 & \textbf{71.74} & 62.78 \\
Nutrition and Food & 42.83 & 18.99 & 21.55 & 49.81 & 23.26 & 67.44 & 67.05 & 69.22 & 75.73 & \textbf{78.56} \\
Other & 48.46 & 19.39 & 23.64 & 52.96 & 24.35 & 68.32 & 58.91 & 67.46 & 71.63 & \textbf{78.10} \\
Professions & 51.53 & 17.88 & 17.55 & 47.18 & 24.35 & 69.65 & 64.62 & 70.24 & 75.03 & \textbf{75.15} \\
Public Places & 54.48 & 18.34 & 21.72 & 60.34 & 24.83 & 77.59 & 66.55 & 75.00 & 82.35 & \textbf{82.41} \\
Reviews & 52.86 & 18.57 & 23.19 & 50.00 & 24.29 & 64.29 & 60.00 & 69.12 & 70.00 & \textbf{72.46} \\
Science & 48.75 & 21.42 & 18.79 & 43.63 & 24.31 & 67.28 & 63.94 & 68.78 & \textbf{74.34} & 71.56 \\
Social Media & 51.34 & 23.47 & 21.52 & 56.72 & 29.58 & 70.42 & 63.88 & 73.27 & \textbf{74.26} & 82.06 \\
Sports & 51.55 & 22.33 & 21.36 & 60.53 & 29.26 & 73.53 & 64.50 & 69.98 & 78.79 & \textbf{81.46} \\
\bottomrule
\end{tabular}
\caption{Mean \texttt{GoEval}-mini for various application domains across multiple VLMs. The best result in each domain is represented in \textbf{BOLD}. Note, this also includes the domains that were excluded from main paper's analysis because of having less than 300 task instances.}
\label{tab:domain_results_all}
\end{table*}

\begin{table*}[htbp]
\centering
\small
\begin{tabular}{p{3.2cm}|p{0.8cm}p{0.8cm}p{0.8cm}p{0.8cm}p{0.8cm}p{0.8cm}p{0.8cm}p{0.8cm}p{0.8cm}p{0.8cm}}
\toprule
\textbf{Knowledge Type} & \textbf{Intern-VL-2-1B} & \textbf{Qwen-2-VL-2B} & \textbf{Pali-Gemma-3B} & \textbf{LLaVA-v1.6-Mistral-7B} & \textbf{Qwen-2-VL-7B} & \textbf{Intern-VL-2-8B} & \textbf{Cog-VLM-2-Llama-3-19B} & \textbf{Gemini-1.5-Flash} & \textbf{Gemini-1.5-Pro} & \textbf{GPT-4o mini} \\
\midrule
Commonsense & 49.95 & 21.10 & 19.88 & 53.33 & 25.25 & 71.06 & 63.91 & 71.24 & 76.32 & \textbf{78.41} \\
Cultural & 48.36 & 18.70 & 15.57 & 49.65 & 23.26 & 70.54 & 66.70 & 70.21 & \textbf{75.77} & 75.31 \\
Geographical & 56.58 & 19.95 & 19.20 & 51.37 & 24.77 & 71.00 & 64.98 & 73.28 & \textbf{78.50} & 78.12 \\
Literary & 44.25 & 17.42 & 13.24 & 41.46 & 20.91 & 68.64 & 71.58 & 67.97 & \textbf{73.78} & 66.78 \\
Mathematical & 38.13 & 11.33 & 12.40 & 22.93 & 16.13 & 61.20 & 60.51 & 54.30 & \textbf{63.67} & 50.88 \\
Other & 41.82 & 13.64 & 20.00 & 45.45 & 16.36 & 62.73 & 57.27 & 63.30 & 67.59 & \textbf{76.85} \\
Scientific & 41.72 & 16.78 & 18.98 & 35.76 & 21.63 & 69.54 & 68.81 & 67.87 & \textbf{74.22} & 66.59 \\
Social & 48.12 & 21.49 & 19.23 & 55.04 & 27.87 & 70.09 & 63.57 & 70.63 & 74.24 & \textbf{77.20} \\
Technical & 46.15 & 17.74 & 15.31 & 42.13 & 24.29 & 69.17 & 61.32 & 64.19 & 71.87 & \textbf{72.10} \\
Temporal & 49.62 & 12.69 & 10.14 & 25.13 & 16.15 & 68.72 & 67.31 & 63.42 & \textbf{69.45} & 58.95 \\
Visual & 52.03 & 21.97 & 19.54 & 52.92 & 24.13 & 68.67 & 60.27 & 67.87 & 73.81 & \textbf{77.82} \\
\bottomrule
\end{tabular}
\caption{Mean \texttt{GoEval}-mini for all knowledge types across multiple VLMs. The best result in each knowledge type is represented in \textbf{BOLD}. Note, this also includes the knowledge types that were excluded from main paper's analysis because of having less than 300 task instances.}
\label{tab:knowledge_types_results_all}
\end{table*}

\end{document}